\documentclass{article}

\usepackage{arxiv}

\usepackage[utf8]{inputenc} 
\usepackage[T1]{fontenc}    
\usepackage{hyperref}       
\usepackage{url}            
\usepackage{booktabs}       
\usepackage{amsfonts}       
\usepackage{nicefrac}       
\usepackage{microtype}      
\usepackage{lipsum}		
\usepackage{graphicx}
\usepackage{natbib}
\usepackage{doi}
\usepackage{amsmath}
\usepackage{multirow}
\usepackage{longtable}

\title{Recent Applications of Machine Learning, Remote Sensing, and IoT Approaches in yield prediction: A Critical Review}

\author{ \href{https://orcid.org/0000-0000-0000-0000}{\includegraphics[scale=0.06]{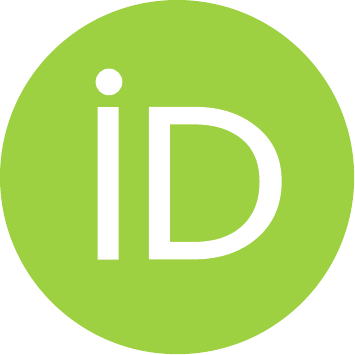}\hspace{1mm}Fatima Zahra Bassine}
  \\
	International Water Research Institute (IWRI)\\
        Mohammed VI Polytechnic University (UM6P)\\
        Lot 660, Hay Moulay Rachid, Benguerir 43150, Morocco \\
	\texttt{fatimazahra.bassine@um6p.ma} \\
	\And
	\href{https://orcid.org/0000-0000-0000-0000}{\includegraphics[scale=0.06]{orcid.pdf}\hspace{1mm}Terence Epule Epule} \\
	Research and Development Unit in Agriculture and Agri-food of Abitibi-Témiscamingue (URDAAT)\\
 Université du Québec en Abitibi-Témiscamingue\\
 79 Côté Street, Notre-Dame-du-Nord, QC J0Z 3B0, Canada \\
	\texttt{terenceepule.epule@uqat.ca} \\
	\AND
	Ayoub Kechchour \\
	   International Water Research Institute (IWRI)\\
        Mohammed VI Polytechnic University (UM6P)\\
        Lot 660, Hay Moulay Rachid, Benguerir 43150, Morocco \\
	\texttt{ayoub.kechchour@um6p.ma} \\
	\And
	Abdelghani Chehbouni \\
	International Water Research Institute (IWRI)\\
        Mohammed VI Polytechnic University (UM6P)\\
        Lot 660, Hay Moulay Rachid, Benguerir 43150, Morocco \\
	\texttt{abdelghani.chehbouni@um6p.ma} \\
}

\renewcommand{\shorttitle}{\textit{arXiv} Template}

\hypersetup{
pdftitle={A template for the arxiv style},
pdfsubject={q-bio.NC, q-bio.QM},
pdfauthor={David S.~Hippocampus, Elias D.~Striatum},
pdfkeywords={First keyword, Second keyword, More},
}

\begin{document}
\maketitle

\begin{abstract}
The integration of remote sensing and machine learning in agriculture is transforming the industry by providing insights and predictions through data analysis. This combination leads to improved yield prediction and water management, resulting in increased efficiency, better yields, and more sustainable agricultural practices. Achieving the United Nations' Sustainable Development Goals, especially "zero hunger," requires the investigation of crop yield and precipitation gaps, which can be accomplished through, the usage of artificial intelligence (AI), machine learning (ML), remote sensing (RS), and the internet of things (IoT). By integrating these technologies, a robust agricultural mobile or web application can be developed, providing farmers and decision-makers with valuable information and tools for improving crop management and increasing efficiency. Several studies have investigated these new technologies and their potential for diverse tasks such as crop monitoring, yield prediction, irrigation management, etc. Through a critical review, this paper reviews relevant articles that have used RS, ML, cloud computing, and IoT in crop yield prediction. It reviews the current state-of-the-art in this field by critically evaluating different machine-learning approaches proposed in the literature for crop yield prediction and water management. It provides insights into how these methods can improve decision-making in agricultural production systems. This work will serve as a compendium for those interested in yield prediction in terms of primary literature but, most importantly, what approaches can be used for real-time and robust prediction.
\end{abstract}


\section{Introduction}
\label{sec:introduction}
\begin{figure}[h!]
    \centering
    \includegraphics[width=0.85\textwidth]{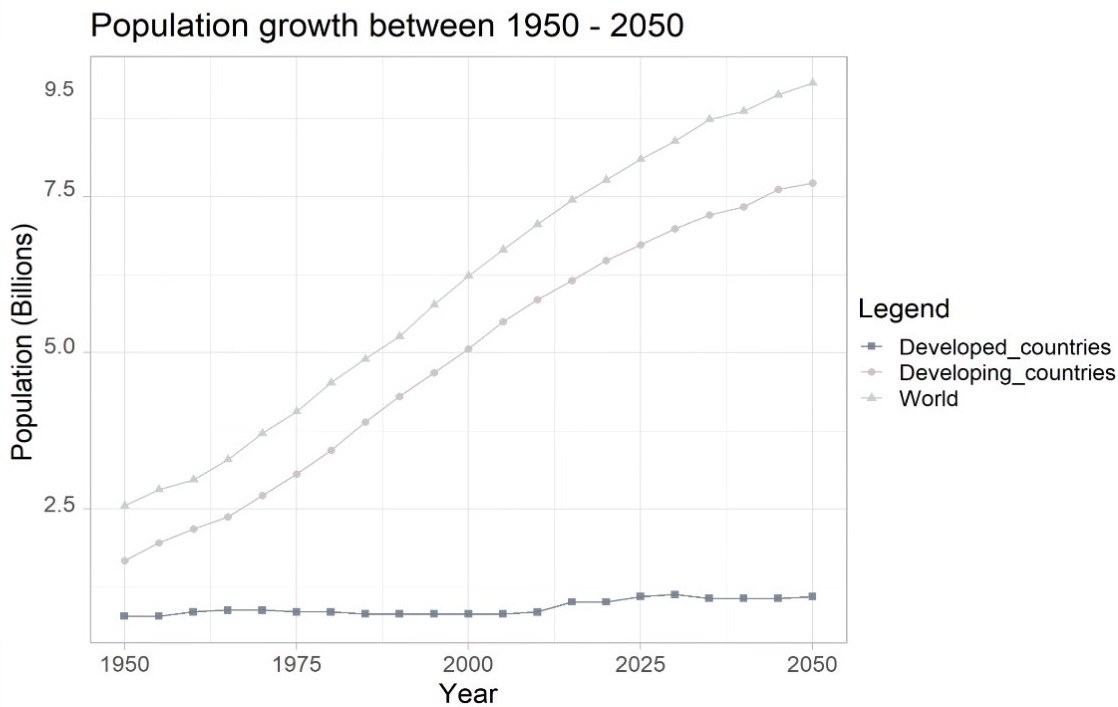}
    \caption{Population growth between 1950 - 2050 Source: Based on Data from \cite{alexandratos2012world}.}
    \label{fig1}
\end{figure}

According to the United Nations, the world population recently attended 8 billion on 15 November 2022 \cite{united_nations_day_2022}. Predictions for the upcoming years are 9.7 billion by 2050 \cite{leridon_world_2020}. With the exponential growth of the global population (Figure \ref{fig1}), the FAO estimates that mankind will need to produce 70\% more food by 2050 \cite{the_food_and_agriculture_organization_fao_2050_2009}. The challenge for the agricultural sector is to meet the growing demand to feed the entire planet, and farmers will have to acquire new tools to increase their production through robust prediction.

\begin{figure}[h!]
    \centering
    \includegraphics[width=0.72\textwidth]{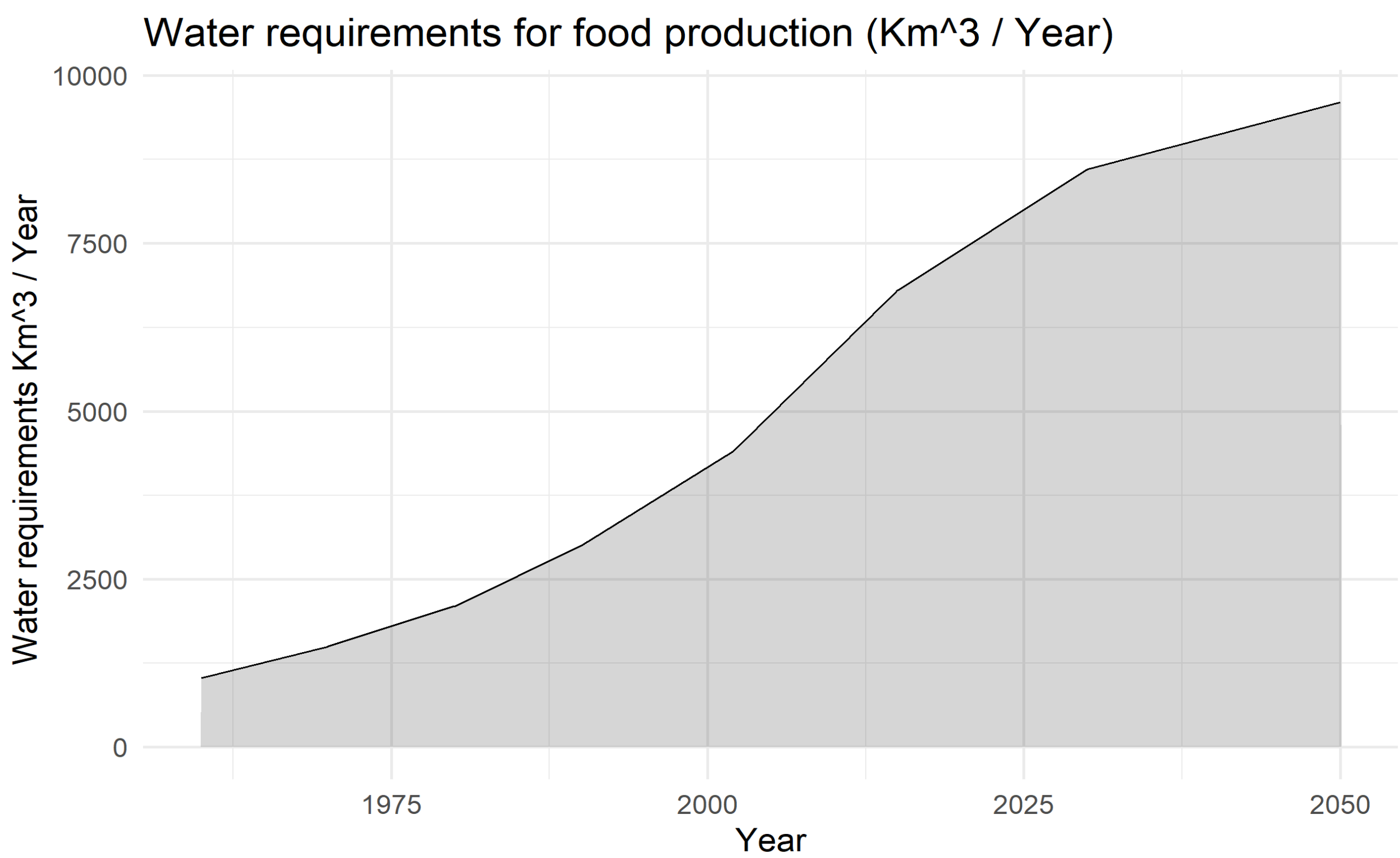}
    \caption{Water requirements for food production ($Km^3/Year$). Source: Authors’ conceptualization based on data from \cite{boretti2019reassessing}.}
    \label{fig2}
\end{figure}

The difficulties of world agriculture were examined in depth and the need for sustainable development of crop yields to ensure food security in a context of increasing pressure on natural resources \cite{mc_carthy_global_2018}. In addition to climate change, the changing structure and decline of the labor force, and arable land are concerns that hinder the agricultural sector \cite{lipper_climate-smart_2014}. To grapple with these challenges, the sector is now consistently driven by innovation in the context of Climate Smart Agriculture (CSA) and is experiencing an accelerated transformation with digital technologies such as AI ML RS, and IoT. These solutions help farmers bridge the gap between supply and demand, improving yields and profitability and encourage sustainable agriculture. Furthermore, the challenges are numerous: climate change, water scarcity (Figure \ref{fig2}), soil degradation, and achieving food security. To meet these challenges, the industry must reinvent itself and move towards what is called Agriculture 4.0, in this context, adopting internet connectivity solutions in agricultural practices has become crucial to reduce the need for manual labor and therefore enhance automation. Nevertheless, the recent progress of innovations, including RS, ML and crop model simulation, has emerged to predict yield more precisely, with the ability to analyze a massive amount of data using improvements in scalability and computing power \cite{sharma_enabling_2022,shahhosseini_forecasting_2020,basso_seasonal_2019}. Multiple research projects are currently demonstrating relatively higher prospects in the use of machine learning algorithms than conventional statistics \cite{elavarasan_forecasting_2018,rehman_current_2019,mosaffa_application_2022}. On such project is the ‘’Pan Moroccan crop precipitation platform-PAMOCPP’’ which seeks to combine several approaches (ML, Statistics, Experimentation, RS and Digital Applications) in yield gap predictions.

\begin{figure}[h!]
    \centering
    \includegraphics[width=\textwidth]{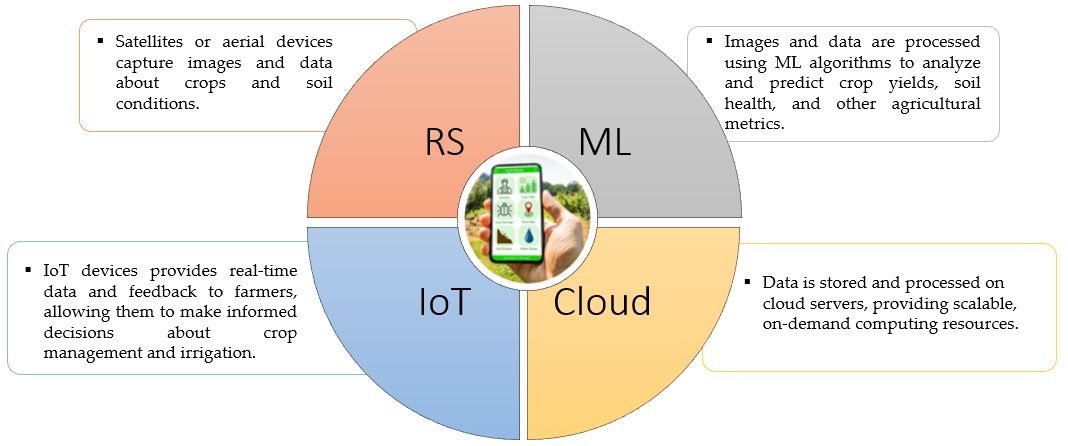}
    \caption{Illustration of key aspects of the review. Source: Authors’ conceptualization.}
    \label{fig3}
\end{figure}

\begin{figure}[h!]
    \centering
    \includegraphics[width=0.95\textwidth]{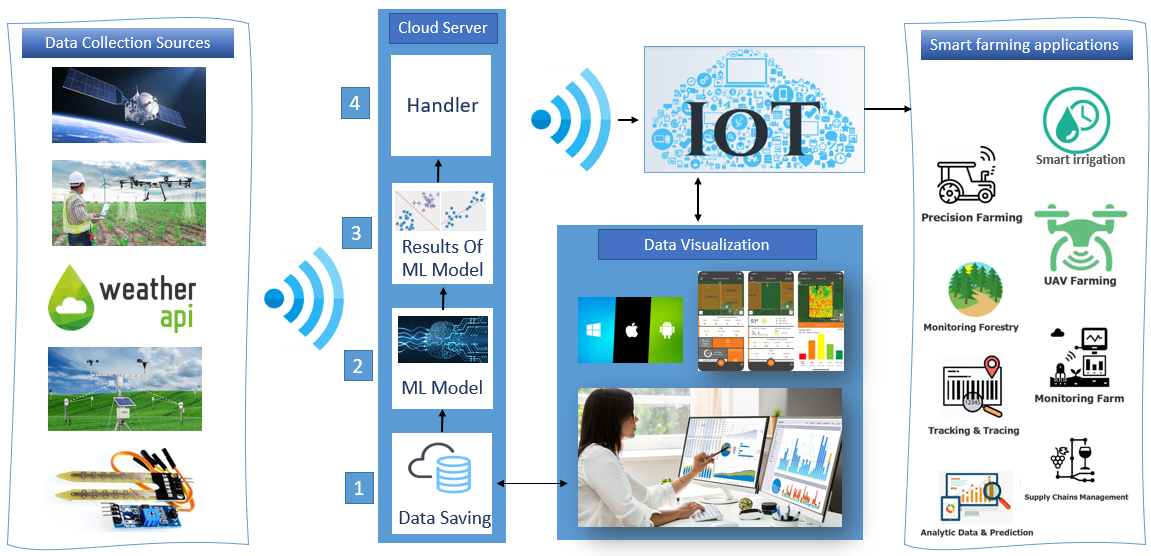}
    \caption{Illustration of IoT Smart Farming Applications based on ML. Source: Authors’ conceptualization.}
    \label{fig4}
\end{figure}

Multiple researchers and publications have reviewed and made models by harnessing the strength of AI and IoT \cite{nwankwo_leveraging_2022,kumar_artificial_nodate,zhang_empowering_2020}. So far, ML has been applied with great success in optimizing irrigation schedules \cite{you_optimizing_2022,guo_optimizing_2021}. In \cite{goldstein_applying_2018,murthy_machine_2019,kim_development_2022,heidari_occupant-centric_2022} showed that using data from weather sensors combined with ML algorithms resulted in a great reduction in water usage without compromising crop yield. Other research papers demonstrated that applying ML techniques to satellite images can improve predictions of yields \cite{khaki_cnn-rnn_2020,pham_enhancing_2022}. These are a few examples of the many methods in which ML is already benefiting agriculture. The combination of RS, ML, Cloud Computing, and IoT play a significant role in agriculture (Figure \ref{fig3}). Remote Sensing captures images and data about crops and soil conditions, which are then processed using ML algorithms to analyze and predict crop yields, soil health, and other agricultural factors. The data is stored and processed on cloud servers, providing scalable and on-demand computing resources. IoT devices provide real-time data and feedback to farmers, allowing them to make informed decisions about crop management and irrigation. The association of these technologies through a mobile application provides a real-time information and tools to farmers to improve crop management and increase efficiency (Figure \ref{fig4}).

This critical review article provides a comprehensive overview of the use of RS and ML algorithms combined with IoTs to predict crop yield and water management and therefore predict decision-making using recent technologies. The article represents a critical review of the current state-of-the-art in forecasting agriculture using RS, ML algorithms and other methods such as IoT for robust yield prediction. This work will serve as major compendium from which researchers and other yield gaps enthusiasts will easily identify which group of methods to use in their yield prediction studies with the goal of understanding the challenges imposed by the various methods. It critically evaluates different machine-learning approaches proposed in the literature for crop yield prediction and water management. It provides insights into how these methods can improve decision-making in agricultural production systems. The paper is organized as follow:

An introduction with an overview of the topic in this first section, the second section focuses on geospatial information especially most used satellites in agriculture and vegetation indices, the third section of the paper reviews different methods that have been proposed for predicting crop yields using ML algorithms. Then the next section discusses common practices especially in water and yield prediction, it also provides a comparative analysis of different algorithms based on their accuracy and performance. Based on their findings, they recommend specific methods that could be used for more accurate predictions. The fifth section provides an overview on IoT, and cloud-based smart farming applications. Then an overview of key observations which presents statistics and critical evaluation of recent existing review articles and identifies some key challenges. Finally, it proposes several practical solutions and recommendations to address agricultural gaps and monitor planetary resources.

\section{Next Generation: Geospatial Information}
\subsection{Satellite basics}
Satellite imagery is a major technological advancement. It offers a considerable number of benefits which makes it an interesting feature to study and exploit \cite{rwanga_accuracy_2017}. Indeed, satellite imagery data have been widely used to inventory and monitor vegetation on a global scale, Thus, remote sensing can provide large-scale mapping and monitoring of crops across a region and field. This information can be used to identify areas with crop stress, areas with high potential for increased yields, and areas needing improvement in irrigation, nutrient management, and pest control. Furthermore, access to timely and accurate information is essential for decision-making. There are many different types of satellites orbiting Earth, but they all have one common goal: to monitor changes in land cover, thus, agricultural applications require specific characteristics in a satellite, such as:

\begin{itemize}
    \item Temporal resolution: The frequency with which image data is obtained from the same point on the surface.
    \item Spatial resolution: The level of detail visible in an image.
    \item Spectral resolution: The width of the spectral bands recorded by the sensor.
    \item Radiometric resolution: Sensitivity of the sensor to capture change.    
\end{itemize}

The most widely used satellites for agricultural applications are presented in Table \ref{table1}:

\begin{table}[h!]
\centering
\caption{Example of widely used satellites and characterization for Spatial, Temporal, and Spectral resolution.}
\label{table1}
\begin{tabular}{llllll} \toprule
\textbf{Characteristic} & \textbf{MODIS} & \textbf{Landsat-9} & \textbf{Sentinel 2} & \textbf{PlanetScope} & \textbf{Rapid eye} \\ \midrule
Temporal\newline Resolution & 1-2 days      & 16 days      & 5 days     & Daily       & Daily     \\ \midrule
Spatial\newline Resolution  & 200, 500, 1000m & 15,30m       & 10,20,60m  & 3meters     & 5meters   \\ \midrule
Spectral\newline Resolution & 36 bands      & 7,8,11 bands & 13 bands   & 4 bands     & 5 bands   \\ \midrule
Cost                & Free          & Free         & Free       &\qquad\$          &\qquad\$ \\ \bottomrule
\end{tabular}
\end{table}

\begin{figure}[h!]
    \centering
    \includegraphics[width=0.83\textwidth]{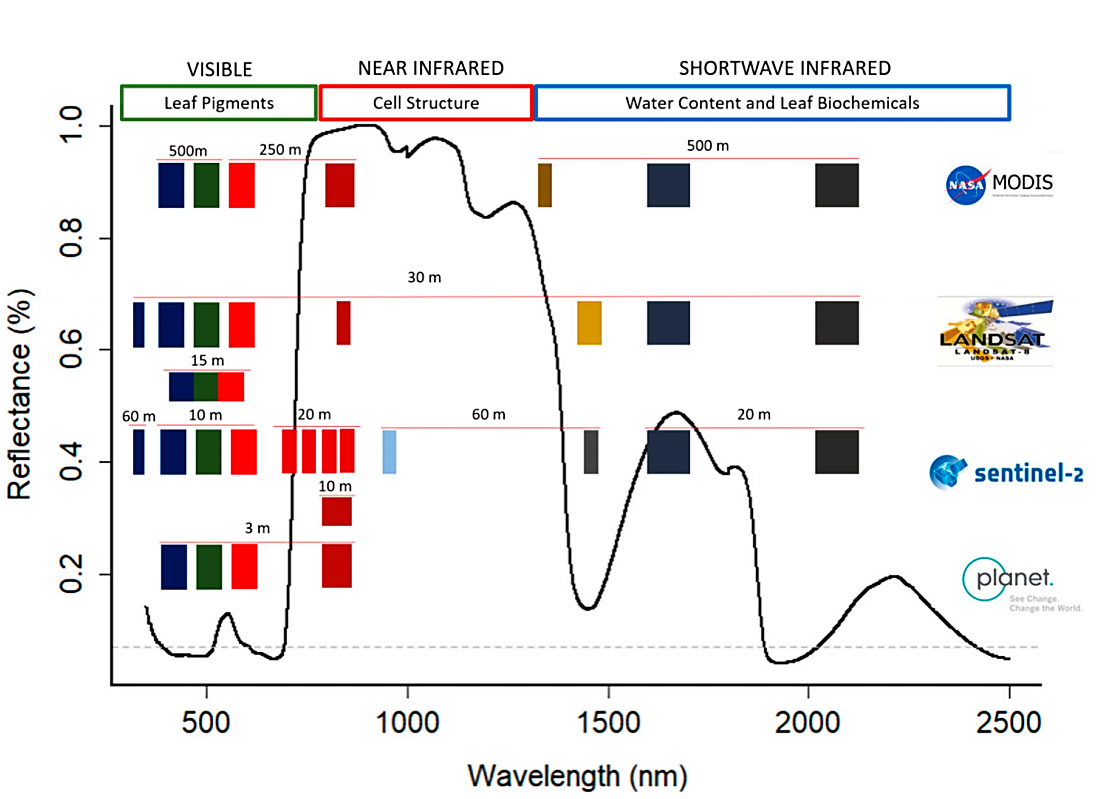}
    \caption{Crop reflectance spectral features and representation of MODIS, Landsat-8, Sentinel-2, and Planet, Source \cite{segarra_remote_2020}.}
    \label{fig5}
\end{figure}

The spectral resolution is a fundamental attribute of satellites because it allows for the precise measurement of the physiological aspects of plants \cite{tucker_satellite_1986,segarra_remote_2020}. There are different types of satellite sensors \cite{nakalembe2021review}, each with its strengths and weaknesses. Table \ref{table1} presents examples of widely used satellite in agriculture. MODIS for example provides excellent temporal resolution but relatively poor spatial resolution for crop monitoring. Landsat-9 provides good spatial resolution but relatively poor temporal resolution. Finally, Planet has very high spatial resolution but only captures three bands in the visible spectrum (red, green, blue) and one in the near infrared (NIR) (Figure \ref{fig5}). Sentinel 2 balances these attributes, providing good temporal and spectral resolutions. However, Sentinel-1 provides valuable information on soil moisture through a technique known as Synthetic Aperture Radar (SAR) interferometry. SAR interferometry works by measuring the phase differences between two or more radar images of the same area, which can be used to estimate soil moisture content, Sentinel-1 has a C-band radar that can penetrate through vegetation canopies and provide insights into the soil moisture beneath \cite{segarra_remote_2020}. The choice of the type of satellite and sensor to use is often based  on factors such as allocation, size of land cover, required precision and accuracy, the threshold in decision-making, the scale of variations, the presence of essential features or management practices, and the frequency of data updates for the study \cite{maurya_remote_2021}.

The Sentinel-2 mission is a part of the Copernicus Program, the European Union's Earth Observation Program \cite{showstack_sentinel_2014}. The main objective of Sentinel-2 is to provide high-resolution optical imagery and multispectral data for land cover characterization and mapping \cite{showstack_sentinel_2014}. In recent years, there has been an increased claim in using Sentinel-2 data for precision agriculture applications \cite{gascon_sentinel-2_2018,bukowiecki_sentinel-2_2021}, and crop management and classification \cite{kumar_crop_2022}. Precision agriculture is a management process that uses temporal and spatial information to support management decisions according to estimated variability for improved resource use efficiency and sustainability of agricultural production, there are several reasons why Sentinel-2 data is well suited for precision agriculture applications as seen in its major ability to track features below the canopies of trees and withing the soil moisture zone. Specifically, some of the benefits are:

\begin{itemize}
    \item Offers high spatial resolution imagery (10 m – 60 m), which can be used to detect small-scale variations in crop conditions.
    \item Provides frequent revisit times (5 days at the equator), allowing farmers to monitor their crops regularly throughout the growing season.
    \item Carries 13 different spectral bands, Sentinel-2 can detect plant stress factors such as water deficit or nutrient deficiencies that may not be visible with other types of satellites.
\end{itemize}

In general, each type of satellite sensor has advantages and disadvantages depending on the specific application \cite{lima_comparing_2019}. However, Sentinel-2 provides an excellent balance between temporal and spatial resolutions while also capturing four narrow band regions in the Red Edge region that can be used to precisely measure plant physiology \cite{segarra_remote_2020-1}. 

In summary, due to its high spatial resolution, frequent revisit times, and multispectral capabilities, Sentinel-2 data is invaluable for supporting precision agriculture management decisions to improve resource use efficiency and sustain agricultural production.

\subsection{Vegetation indices}
Vegetation indices (VIs) is one of the most promising approaches in this area, as it allows for a comprehensive assessment of plant development and growth \cite{segarra_remote_2020-1}. VIs are mathematical combinations of certain spectral bands which allow us to monitor changes in vegetation. Typically, they are derived from satellite or drone reflectance measurements in different spectral bands such as Normalized Difference Vegetation (NDVI), Soil Adjusted Vegetation Index (SAVI), Modified Soil-Adjusted Vegetation Index (MSAVI), Normalized Difference Water Index (NDWI), Enhanced Vegetation Index (EVI), The Chlorophyll Index (CI), Normalized Difference Red Edge index( NDRE), etc. as shown in Table \ref{table2}. VIs provides information on various aspects of plant physiology \cite{gitelson_wide_2004}, including photosynthetic activity \cite{huete_vegetation_2012}, and water status \cite{pocas_predicting_2015,ballester_monitoring_2019}. This makes them an invaluable tool for crop monitoring and yield prediction \cite{panda_application_2010,satir_crop_2016,kureel_modelling_2022}. For example \cite{pei_application_2021} used normalized NDVI for the detection of extreme precipitation change, and \cite{ji_prediction_2021} used VIs to Predict of crop yield.

\begin{table}[h!]
\centering
\caption{Selection of the most relevant VIs for agriculture.}
\label{table2}
\begin{tabular}{p{0.7in}lcp{1.4in}} \toprule
\textbf{Ref.} & \textbf{VI’s} & \textbf{Equation} & \textbf{Examples of application} \\ \midrule
\cite{pettorelli_using_2005} & NDVI & $\dfrac{NIR-Red}{NIR+Red}$ & \begin{tabular}{@{}p{1.4in}@{}}
Estimating the state of plant health based on how the plant reflects light at frequencies, helps agronomists recognize stressed crops up to 2 weeks before the naked eye can see. Since crop stress is better apparent in the near-infrared light spectrum.
\end{tabular} \\ \midrule

\cite{huete_soil-adjusted_1988} & SAVI & $\dfrac{(NIR-Red)}{(NIR+Red+L)} \times (1+L)$ & \begin{tabular}{@{}p{1.4in}@{}}
Beneficial for crops with more visible soils \ Early growing stages.
\end{tabular} \\ \midrule

\cite{qi_modified_1994} & MSAVI & $\dfrac{2 \times NIR+1-\sqrt{(2 \times NIR+1)^2-8 \times (NIR-Red)}}{(NIR+Red+L)}$ & \begin{tabular}{@{}p{1.4in}@{}}
Used in remote sensing to detect irregular seed growth. It can be approximated to weather data on the graph, indicating the correlation between extreme weather and crop health.
\end{tabular} \\ \midrule

\cite{gao_ndwinormalized_1996} & NDWI & $\dfrac{Green-NIR}{Green+NIR}$ & \begin{tabular}{@{}p{1.4in}@{}}
Estimates the relative water content in leaves. Provides a prominent indicator of water stress level in the crop.
\end{tabular} \\ \midrule

\cite{gurung_predicting_2009} & EVI & $8 \times \dfrac{NIR-Red}{(NIR+C1 \times Red-C2 \times Blue+L)}$ & \begin{tabular}{@{}p{1.4in}@{}}
Used for crops with high canopy density, NDVI tends to saturate while EVI is more robust to use for crops or growing stages.
\end{tabular}  \\ \midrule

\cite{wu_remote_2009} & CI & $\dfrac{NIR}{Green-1}$ & \begin{tabular}{@{}p{1.4in}@{}}
Assesses the level of chlorophyll in the plant, indicative of nutrient requirements and yield.
\end{tabular} \\ \midrule

\cite{eitel_broadband_2011} & NDRE & $\dfrac{NIR-RedEdge}{NIR+RedEdge}$ & \begin{tabular}{@{}p{1.4in}@{}}
Monitor the nitrogen stress in the crop and create prescription maps.
\end{tabular} \\ \bottomrule
\end{tabular}
\end{table}

These indices provide information on the current physical state of plants, including plant color, density, and water content. We can categorize these indices into two main categories: Crop activity indices which provides details on the physical state of plants such as SAVI, EVI, LAI, and NDWI and crop productivity to estimate yields and provide information about biochemical requirements, for example, CHI, NDRE, and Near-Infrared Reflectance of Vegetation (NIRv). There are numerous advantages to using vegetation indices in remote sensing applications, They offer a high degree of data precision due to the large number of spectral bands, vegetation indices can be associated with additional types of data such as weather data \cite{campos_mapping_2019,li_hierarchical_2020,ruan_integrating_2022}, historical yield data \cite{anghileri_comparison_2022,wimberly_cloud-based_2022} to create more comprehensive models that forecast crop yields with high accuracy. Lee et al. \cite{lee_predicting_2020} presents a novel approach employing NIRv to predict the yield of major crops at the state scale. The proposed approach can also forecast the crop yield more easily and promptly than extant methods based on meteorological data and processing models; this study suggests that NIRv has the potential for robust estimation of crop yield. Furthermore, NDVI is a powerful tool for mapping and monitoring vegetation because of its ease of computation, capacity to offer vegetation information, reliable findings, and interaction with other data sources. Because of these benefits, NDVI is a powerful tool for a variety of applications in agriculture, forestry, and land use management. NDVI has significant drawbacks, despite its advantages \cite{xue_significant_2017}. NDVI is sensitive to soil and atmospheric conditions, leading to variability in results and difficulty in interpretation, NDVI is ineffective in places with low vegetation density or with little or no plant cover, making mapping and monitoring difficult \cite{huang2021commentary}. These limitations should be taken into consideration when using NDVI for vegetation mapping and monitoring. SAVI, on the other hand, is improved version of NDVI that offers better soil correction and performance in locations with changing soil background \cite{huete_soil-adjusted_1988}. It is, however, a more complex index that may not deliver considerable benefits in all circumstances \cite{baret1991potentials}. NDWI is a useful index for mapping and monitoring open water bodies and wet vegetation because of its improved precision, sensitivity to moisture, and capacity for integration with other data sources \cite{gao_ndwinormalized_1996}. However, when utilizing NDWI for mapping and monitoring water bodies, its limitations in urban and densely vegetated regions, should be considered \cite{ismail2022application}. EVI is an important index for mapping and monitoring vegetation cover and crop yield due to its improved sensitivity, correction for atmospheric and canopy effects, and improved performance in high latitude regions \cite{jiang2008development}. However, its limitations in dry regions and increased computational requirements should be taken into consideration when using EVI for mapping and crop monitoring \cite{ben2006assessing}.

In summary the combination of satellite imagery and data from Sentinel 2 through the use of VI’s provides a powerful tool for multiple application in modern agriculture Table \ref{table3}. The high-resolution images and data allow farmers to monitor their crops and make better decisions about irrigation, fertilization, etc. This information can help farmers increase yields while reducing inputs cost.

\begin{table}[h!]
\centering
\caption{Examples of Sentinel-2 data for precision agriculture applications.}
\label{table3}
\begin{tabular}{p{1.2in}lp{3.8in}} \toprule
\textbf{Ref} & \textbf{Application} & \textbf{Method} \\ \midrule
\cite{hunt2019high,fieuzal2020combined,habyarimana2019towards} & Yield Prediction & VIs and yields, jointly with climatological data for dataset setup. \newline
Regressions, random forest, and ML to forecast yields. \newline
Coupling with crop functioning models, FAPAR, LAI, SLA, and light for efficiency. \\ \midrule

\cite{castaldi2019evaluating} & Soils & Determining Soil organic matter with VIs and Sentinel-2 bands. \newline
Classification of soil degradation. \newline
The wavelength of the organic matter spectral feature in the visible is near to S2 red band. \\ \midrule

\cite{defourny2019near,son2020classification} & Fields monitoring & VIs for cropping practices assessment; regional and nation-scale cropland and crop type classification with S2, time series and retrieval of biophysical and vegetation radiometric indexes. \\ \midrule

\cite{vanino2018capability,rozenstein2018estimating} & Water & Sentinel-2 vegetation parameters, surface, and crop height for FAO-56.\newline
Penman-Monteith ET estimation; red and red-edge bands to predict crop coefficients (Kc). Irrigated and rain-fed cropland differentiation.\newline
Combination of Sentinel-2 and Aqua crop. \\ \midrule
\end{tabular}
\end{table}

\section{Machine Learning Applications in Smart Agriculture}
\begin{figure}[h!]
    \centering
    \includegraphics[width=\textwidth]{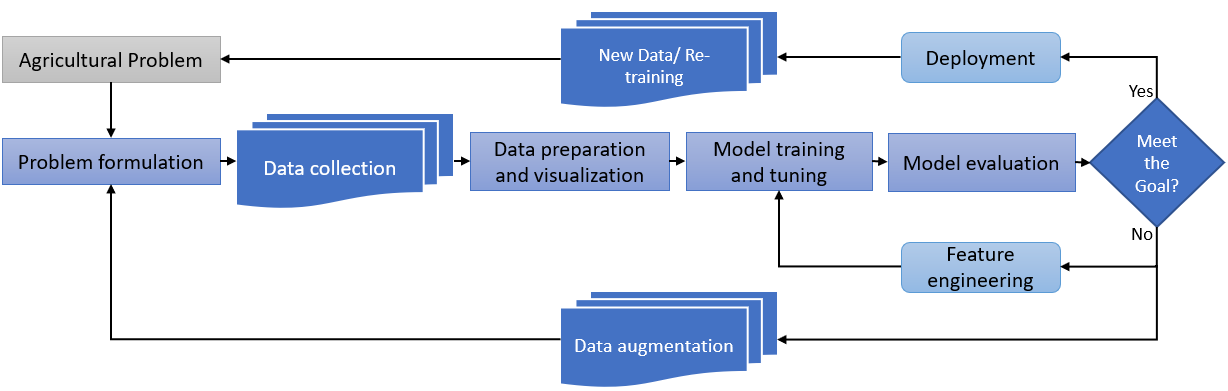}
    \caption{Machine Learning Pipeline.}
    \label{fig6}
\end{figure}

Digital transformation is a central element of efficient management. Today’s technology offers a wide range of powerful tools throughout the agricultural and agrifood value chain to improve productivity, quality, and competitiveness and, consequently, constitute a growth relay for the agricultural sector. The collection and analysis of big data in agriculture will represent a significant value in the future of modem farming in preserving ecosystems \cite{singh_geoinformatics_2022}. ML is one of the key technologies used to process agricultural data, they are used for predictive modeling, furthermore the use of ML depends on specific agricultural problem and the complexity of data (Figure \ref{fig6}). In addition to ML, IoT technology offers more benefits in real life. Researchers are investigating this technology for broader use and maximum profit. IoT devices have been used on farms for some time now \cite{phasinam_applicability_2022}, the sensors attached to these devices collect extensive amounts of data that can be processed to indicate patterns and trends that would otherwise be invisible. This has led to significant improvements in yields and efficiency on many farms around the world. With better data analysis, farmers can make even finer adjustments to their operations, leading to further productivity and reducing wastage (of both resources and money). In addition, by collecting data on devices such as weather patterns and soil conditions over time, farmers can build up detailed models that can help them predict problems before they arise \cite{akhter_precision_2022,koirala_deep_2019}.

\subsection{Machine Learning Approaches}
\begin{figure}[h!]
    \centering
    \includegraphics[width=\textwidth]{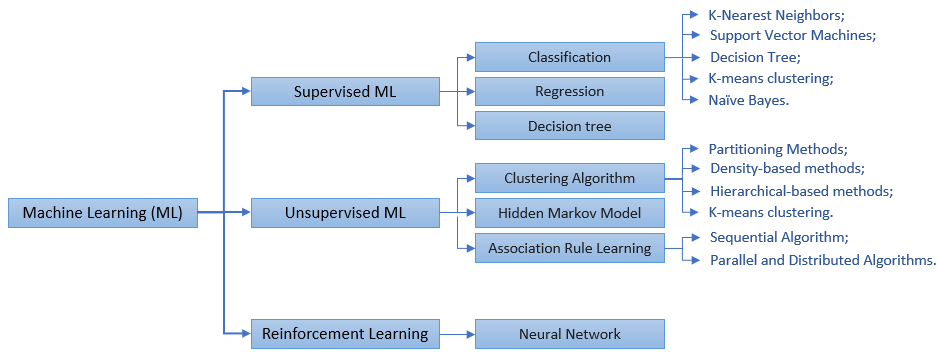}
    \caption{Machine Learning Approaches.}
    \label{fig7}
\end{figure}

ML is a field of AI that employs algorithms to learn from data. There are three major types of ML: supervised, unsupervised, and reinforcement learning (Figure \ref{fig7}). Most machine-learning techniques used in remote sensing are either supervised or unsupervised. Supervised methods require ground data to be used for training, while unsupervised methods don't. Both approaches have their advantages and weaknesses, which will be discussed in more detail below.

\begin{figure}[h!]
    \centering
    \includegraphics[width=\textwidth]{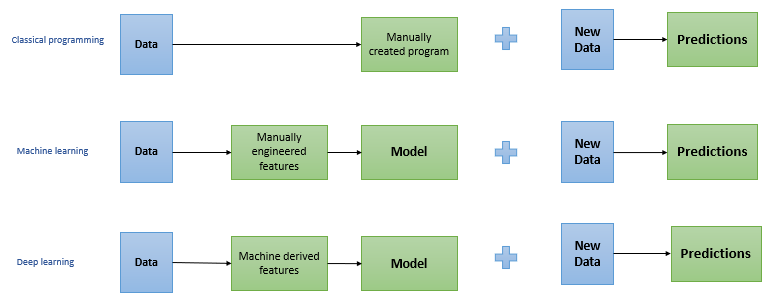}
    \caption{Classical programming Vs Machine learning Vs Deep learning.}
    \label{fig8}
\end{figure}

Supervised methods are often more accurate than unsupervised ones, as they can learn from known examples and correct any mistakes, Bisgin et al. \cite{bisgin_comparing_2018}  presented a new SVM-based technique model, which improved the average accuracy by up to 85\%. in comparison with the ANN model 80\% accuracy after considerable parameter optimization. Although unsupervised learning algorithms are powerful tools that can be used to identify patterns from data, they require less labeled data than supervised learning algorithms and can often find hidden patterns in data. Additionally, unsupervised learning can be used on various data types, one major limitation is that unsupervised learning algorithms often produce results that are hard to interpret. In traditional ML (Figure \ref{fig8}), the algorithm is given a set of relevant features to analyze. However, in deep learning, the algorithm is given raw data and decides what features are relevant but requires GPUs (graphics processing units) to perform high-power computation on complex architectures \cite{magalhaes_benchmarking_2023}, they require many training examples to learn complex patterns from given data, for example, different common wireless nodes and remote sensors capturing environmental parameters related to crops in agriculture \cite{geng_agricultural_2017}, these models can then be operated to make predictions on new data.

These ML algorithms can optimize irrigation schedules to reduce water usage without compromising crop yield by analyzing weather patterns and plant growth \cite{linaza2021data}. A novel deep learning method for automatic detection and mapping of center pivot irrigation systems using satellite data, this method helps providing an estimation of the area equipped with irrigation systems and also helps monitoring and estimating freshwater use and demand \cite{saraiva_automatic_2020}. And helps to develop precision farming techniques that target specific field areas with precise amounts of fertilizer or pesticides \cite{khan2021deep}. In general ML algorithms helps farmers predict, monitor, and use resources more efficiently while minimizing environmental impact. This information can help farmers take proactive measures to prevent problems before they cause significant damage. 

\subsection{Comparative study on classic machine learning algorithms.}
In recent years, ML algorithms have been used increasingly for crop yield prediction in agriculture. Many different ML algorithms are used for this goal, each with advantages and weaknesses (Table \ref{table4}). This section will discuss the most used ML algorithms for crop yield prediction.

\begin{table}[h!]
\centering
\caption{Comparison of widely used prediction algorithms used in agriculture.}
\label{table4}
\begin{tabular}{lp{1.7in}p{1.7in}p{1.7in}} \toprule
\textbf{Algorithm} & \textbf{Objectives} & \textbf{Advantages} & \textbf{Limitations} \\ \midrule
SVM & Used for both classification and regression tasks. The main idea behind SVMs is to find a hyperplane that maximally separates the data points of one class from those of the other. & Handles nonlinear relationships. That is important in yield prediction because weather patterns and other environmental factors often have nonlinear relationships with crop yields \cite{chen_assessing_2016}.\newline
Deals with high dimensional data sets, which are common in yield prediction \cite{shi_research_2011} and water management applications \cite{deka_support_2014}. & Sensitive to outliers in the data set. This means that supervision must be taken to ensure that the data used for training the model are representative of the real conditions that will be encountered during the application. \\ \midrule

RF & Integrates multiple decision trees and outputs the most frequent or average prediction from those trees.\newline
Used for classification and regression tasks due to their high accuracy and robustness against overfitting. & Relatively easy to tune and does not require extensive data preprocessing like some other ML algorithms 
\cite{chen_mapping_2021}. & Computationally expensive when working with large datasets.\newline
Needs more time for training as it integrates a lot of decision trees to determine the class. \\ \midrule

ANN & Creates an AI system capable of making predictions or decisions without humans' external programming or instruction. & Ability to learn from data, recognize patterns, and generalize from examples, handle noisy data; and deals with nonlinear problems. & Complex structure. \newline
Can be easily overfit on training data if it is not carefully designed. 
Requires a large amount of training data to learn effectively. \\ \midrule

DNN & Helps generate a flexible and effective detection approach. & Used for both classification and regression; Automatically learns high level features, handles complex data with high accuracy. & Requires large amounts of data to train the model, which can be difficult to obtain in some cases.\newline
Can be computationally intensive, making it impractical for real-time applications. \\ \bottomrule
\end{tabular}
\end{table}

\clearpage

\subsection{Performances of Machine Learning Algorithms in Predicting the Productivity of Conservation Agriculture at a Global Scale}
Evaluation metrics can define a model's performance. Evaluation measures play a significant role because they can differentiate among the outcomes of different learning models. There are various performance metrics applied to evaluate the performance of regression models (Table \ref{table5}), including mean absolute error (MAE), mean squared error (MSE), root mean square error (RMSE), determination coefficient (R-squared) and mean absolute percentage error (MAPE). In \cite{salih2022evaluation} authors used RMSE, R and R$^2$ to compare and evaluate the accuracy of satellite data and multiple weather station datasets in estimating precipitation in the Tensift basin. The study compares various satellite products with the observed precipitation data from six regional weather stations. The purpose is to determine how the satellite data aligns with the observed precipitation data from the weather stations. Each of these evaluation measures has strengths and weaknesses, so it is vital to choose the right metric for the job. MAE is a good choice if we want to know how close our predictions are to the actual values, while MSE is better suited for measuring the overall accuracy of our predictions. RMSE considers both factors, therefore, often used as a general measure of predictive power. R-squared tells us how much variation in our dependent variable can be explained by our independent variables. At the same time, MAPE gives us an idea of how accurate our predictions are compared to those made by other models. In short, choosing an appropriate evaluation metric is essential for accurately assessing a model's performance. One or more of these measures may be more valuable than others, depending on what we want to know about our model's predictive efficiency.

\begin{table}[h!]
\centering
\caption{Most used regression metrics.}
\label{table5}
\begin{tabular}{llp{1.9in}c} \toprule
\textbf{Metric} & \textbf{Space} & \textbf{Definition} & \textbf{Formula} \\ \midrule
MAE & $\geq 0$ & \begin{tabular}{@{}p{1.9in}}
\textit{Mean absolute error} represents the difference between the original and predicted values extracted by averaged the absolute difference over the data set.
\end{tabular} & $\displaystyle\dfrac{1}{N} \sum_{i=1}^N \left|y_i - \hat{y}\right|$ \\ \midrule

MSE & $\geq 0$ & \begin{tabular}{@{}p{1.9in}}
\textit{Mean Squared Error} represents the difference between the original and predicted values extracted by squared the average difference over the data set.
\end{tabular} & $\displaystyle \frac{1}{N} \sum_{i=1}^N (y_i-\hat{y})^2$ \\ \midrule

RMSE & $\geq 0$ & \begin{tabular}{@{}p{1.9in}}
\textit{Root Mean Squared Error} is the error rate by the square root of MSE.
\end{tabular} & $\displaystyle\sqrt{MSE} = \sqrt{\frac{1}{N} \sum_{i=1}^N (y_i-\hat{y})^2}$ \\ \midrule

RRMSE & \% & \begin{tabular}{@{}p{1.9in}}
\textit{Relative Root Mean Squared Error} is the RMSE normalized by the mean of the $y$.
\end{tabular} & $\dfrac{RMSE}{\bar{y}}$ \\ \midrule

R-squared & [0-1] & \begin{tabular}{@{}p{1.9in}}
Coefficient of determination represents the coefficient of how well the values fit compared to the original values. The value from 0 to 1 interpreted as percentages. The higher the value is, the better the model is.
\end{tabular} & $\displaystyle 1-\frac{\sum_{i=1}^N (y_i-\hat{y})^2}{\sum_{i=1}^N (y_i-\bar{y})^2}$ \\ \bottomrule

\multicolumn{4}{l}{Where $\hat{y}$ is the predicted value of $y$ and $\bar{y}$ is the mean value of actual values $y$.}
\end{tabular}
\end{table}

\clearpage

Classification metrics are a set of mathematical formulas used to evaluate the performance of a ML model. These metrics measure how accurately a model can classify data into specific categories or classes and provide insight into its effectiveness. The most used classification metrics include precision, recall/sensitivity, and F1 score Table \ref{table6}.

\begin{table}[h!]
\centering
\caption{Classification Metrics.}
\label{table6}
\begin{tabular}{lp{1.4in}c} \toprule
\textbf{Metric} & \textbf{Definition} & \textbf{Formula} \\ \midrule
Precision & \begin{tabular}{@{}p{1.4in}}
Measures how accurate and reliable an algorithm's predictions are by considering correct and incorrect classifications TP, FP. A high precision value indicates that the algorithm made few misclassifications while making predictions.
\end{tabular} & $\dfrac{TP}{TP+FP}$ \\ \midrule

Recall/Sensitivity & \begin{tabular}{@{}p{1.4in}}
Measures the proportion out of all actual positives correctly identified by our ML models.\newline
Helps understand if the algorithms have left any true cases during classification, potential opportunities for improvement.
\end{tabular} & $\dfrac{TP}{TP+FN}$ \\ \midrule

F1 score & \begin{tabular}{@{}p{1.4in}}
Allow getting an overall understanding of how well our models perform on average.
\end{tabular} & $\dfrac{2 \times Precision \times Recall}{Precision+Recall}$ \\ \midrule

Accuracy & \begin{tabular}{@{}p{1.4in}}
Measures the percentage of correctly classified classes out of all the instances in a dataset. A high accuracy score indicates that most items have been accurately labeled.
\end{tabular} & $\dfrac{TP+TN}{TP+FP+TN+FN}$ \\ \midrule

Kappa & \begin{tabular}{@{}p{1.4in}}
Evaluate the agreement between different annotators or between a human annotator and a machine classifier. They help to measure the reliability and consistency of image annotations and are particularly useful when there is significant class imbalance or when the classes are difficult to distinguish.
\end{tabular} & 
$\begin{array}{c}
\dfrac{\text{Observed agreement} - \text{Expected agreement}}{1 - \text{Expected agreement}} \\
\text{Observed agreement} : \dfrac{TP + TN}{TP + TN + FP + FN} \\
\text{Expected agreement}:\\
 \dfrac{((TP + FP) \times (TP + FN) + (TN + FP) \times (TN + FN))}{(TP + TN + FP + FN)^2}
\end{array}$ \\ \bottomrule
\multicolumn{3}{l}{Where TP, FP, TN, and FN respectively represent False Positive, False Negative, True Positive, and True Negative.}
\end{tabular}
\end{table}

The performance of ML algorithms in predicting the productivity of agriculture has been extensively studied in the literature \cite{mupangwa2020evaluating,ps2019performance,su2022performances,bakthavatchalam2022iot}. Several studies have demonstrated the potential of using ML techniques to predict the productivity of conservation agriculture practices. For example, \cite{galphade2022crop} used a decision tree algorithm to predict the productivity in smallholder farms. The algorithm was trained on data collected from weather data and NDVI and was able to accurately predict productivity with an accuracy of over 91.5\%. The study concluded that decision tree algorithms have the potential to be used as a tool for predicting the productivity of conservation agriculture. Kaur, et al. \cite{kaur2022trustworthy} found that several studies have demonstrated the potential of using ML techniques, such as decision tree algorithms and support vector machine models, to accurately predict the productivity of conservation agriculture in the region. The study concluded that further research is needed to optimize the use of ML approaches for predicting the productivity of conservation agriculture in Punjab. \cite{aworka2022agricultural} introduced reliable models for agricultural predictions in East Africa using Gradient Boost, Random Forest, and SVM. The RF model, in this study the RF model showed a better result for the crop data with an $R^2$ of 92.272\%, while SVM and GBM achieved accuracies of 90.186\% and 86.377\%, respectively.

These studies demonstrate the promising potential of ML algorithms in predicting the productivity of conservation agriculture at a global scale and highlight the need for further research in this area.

\section{Common practices}
In modern agriculture, sensors, drones, and high-tech technologies quickly evolve into the new norm. Several modem farmers have already adopted smart farming, and its use is increasing and becoming increasingly common among the new generation of educated young farmers. Non-functional Requirements: The non-functional requirements for this product consider availability, reliability, security/privacy, and scalability. The functional requirements indicate the functions and services of the present system. Therefore, water management and yield prediction are the most used practices in agriculture today. With climate change and population growth, it is becoming increasingly important to manage yield and water resources and efficiently. ML has been applied to these practices by using predictive analytics.

\subsection{Yield gaps}
Yield gap is the dissimilarity between the potential yield of a crop and the actual yield achieved under current growing conditions and management practices. It measures the potential for increasing crop productivity. Therefore, achieving food security is a complex challenge that relies on many factors, including the availability of land and water resources, technology, infrastructure, and policies. One crucial factor is crop yield—the amount of food produced per unit of land. Increasing crop yield can help close the gap between actual and potential yield (the maximum possible yield under ideal conditions), reducing hunger and improving food security. There are different ways to measure and model yield at various levels (actual, attainable, potential) and different spatial (field, sub-national, national, and global) and temporal resolution (short-term, long-term). Yield gap analysis is a tool to identify the causes of gaps between actual yields and attainable or potential yields. This information can be used to develop management options to reduce these gaps. Policies encouraging the adoption of technologies that have been shown previously to improve crop yields can also help close productivity gaps.

Finally, practical solutions are needed to close yield gaps in both small and large-scale cropping strategies worldwide to drive advancement in this direction:

\begin{itemize}
    \item Descriptions and strategies to measure and model yield at various levels (actual, attainable, potential) and further scales in length (field, region, province, national, and international) and time (short, long period). 
    \item Designation of the causalities of gaps between yield levels.
    \item Managing options to decrease the gaps.
    \item Approaches to favor the adoption of gap-closing new technologies.
\end{itemize}

Smart farming technologies, such as ML and the IoT, can help to close these gaps by providing real-time data on crop growth and environmental conditions. This information can be used to optimize irrigation, fertilization, and pest management practices, leading to higher yields and more efficient use of resources.

Mueller et al. \cite{mueller_closing_2012} developed the concept of sustainable intensification that aims to increase yields while decreasing environmental impacts. The research highlights the possibility of significantly increasing crop production by 45-70\% through better management of fertilizer, irrigation, and climate, by closing the gaps in yield through improved agricultural practices. Additionally, there are opportunities to reduce the environmental impact of agriculture through better nutrient management while still allowing a 30\% increase in production of major cereals. Jägermeyr et al. \cite{jagermeyr_integrated_2016} reports that simple solutions for small-scale farmers in areas with limited water resources have the potential to significantly increase rainfed crop yields. If implemented, the combined water management strategies discussed in the study could boost global production by 41\% and close the water-related yield gap by 62\%. Although climate change will negatively impact crop yields in many areas, these water management improvements could help mitigate these effects to a large extent. In \cite{rosa_closing_2018} the study showed that the global use of water for irrigation could increase sustainably by 48\%, if it is extended to 26\% of the currently rain-fed agricultural lands. This could result in the production of 37\% more calories, enough to feed an additional 2.8 billion people. Even if the current unsustainable water practices are discontinued, there is still a potential for a 24\% increase in calorie production through sustainable irrigation expansion and intensification. This study \cite{tseng_field-level_2021} focused on measuring the yield gap in rice production in Uruguay, South America and found that the average gap was 19\% in non-hybrid fields and 17\% in hybrid fields. The two most important factors that impacted the yield gap were seed planting date and the rate of nitrogen fertilizer used. The study suggests that further optimization of planting and crop rotation could help reduce the yield gap in some fields. The approach used in this study is data-driven, low-cost and utilizes field-level records to improve productivity.

\subsection{Water management}
Water management is a critical aspect of agriculture. Farmers need to predict how much water their crops will need to avoid wasting resources or putting their plants at risk by giving them too little or too much moisture. Precision agriculture can help with this by using sensors and other technology to collect data about soil moisture levels, weather patterns, and crop growth. This information can make informed decisions about irrigation schedules. Recently, ML has been widely used as a decisive tool to solve crises in the water environment because it can be used to manage, predict, and optimize water resources \cite{allen-dumas_toward_2021,kamarudin_deep_2021}. Accurately forecasting water demand is a promising approach for effectively allocating available water resources. The application of ML can help solve the imbalance in water-supply systems Table \ref{table7}.

\begin{table}[h!]
\centering
\caption{Overview on recent machine learning applications in water management.}
\label{table7}
\setlength{\tabcolsep}{0.02in}
\begin{tabular}{p{0.8in}p{1in}p{1.2in}p{1.1in}p{1in}p{1.1in}} \toprule
\textbf{Ref.} & \textbf{Application} & \textbf{Feature} & \textbf{ML model} & \textbf{Performance} & \textbf{Limitation} \\ \midrule
\cite{reddy2020iot} & Real time smart irrigation system & Temperature,\newline 
Soil moisture, 
Humidity. & DT & Accuracy: based on the size and quality of the data. & Real time data availability and quality. \\ \midrule

\cite{mehdizadeh_comparison_2017} & Estimation of monthly Evapotranspiration in arid and semi-arid regions & Meteorological data,\newline
Temperature,\newline
Humidity,
Wind speed. & SVM-RBF/
MARS & MAE: 0.05\newline
RMSE: 0.07\newline
R: 0.9999 & Need to add different combinations of air temperature parameters to enhance the performance. \\ \midrule

\cite{goap2018iot} & Irrigation requirements predictions & UV Sensor,\newline 
Temperature,\newline 
Soil Moisture,\newline 
Humidity. & K-means SVR & R: 0.55\newline
MSE: 0.13 & Data Security: uses open standard technologies.\newline
Not customized for application-specific scenarios. \\ \midrule

\cite{dimitriadis_applying_2008} & Management of naturel resources in faming land. & Agricultural data & OneR, ZeroR,\newline
NaïveBayes Simple, J48 DecisionStump. & OneR: 91.59 \% & Require full control over all parameters (light, humidity, irrigation, fertilization). \\ \midrule 

\cite{sun_reconstruction_2021} & Global hydrological model: Prediction and data gap filling & ERA5-Land 
(Precipitation, Temperature).
Evapotranspiration; NDVI Vegetation Index; & Automated ML:\newline
CNN; MLR;\newline MLP; ARX–;\newline SARIMAX–ARIM;\newline XGBoost–eXtreme. & NSE: 0.85\newline
R: 0.95\newline
RMSE: 0.09 & Complexity and computational, memory usage. \\ \midrule

\cite{mohammadi_extreme_2015} & Daily dewpoint temperature. & Meteorological data (humidity,air temperature, 
solar radiation and vapor pressure) & ELM & MABE: 0.3240 °C,\newline RMSE:0.5662 °C,
\newline and R: 0.9933 & Missing and unreliable data, it must consider that case studies have different climate conditions. \\ \bottomrule
\end{tabular}
\end{table}

Ghiassi et al. \cite{ghiassi_large_2017} used a dynamic artificial neural network (DANN), a focused time-delay neural network (FTDNN), and The k-nearest neighbors KNN, to predict the daily, weekly, and monthly water demands. Among the three models, DANN displayed the most promising performance. The prediction accuracies of the daily, weekly, and monthly models were 96\%, 99\%, and 98\%, respectively.

Irrigation prediction approach to efficiently manage intelligent automatic irrigation:
\begin{itemize}
    \item Installing the sensors (soil moisture, temperature, and rain).
    \item Linking the set of sensors to an acquisition system.
    \item Using the Node-RED platform, whose objective is to facilitate supervision, storage, and notification.
    \item Processing the collected data using many algorithms: KNN, neural networks, support vector machine, Naive Bayes, and Logistic Regression. The results showed that the KNN algorithm has a better decision-making rate compared to the others, with a rate of 98.3\%.
\end{itemize}

\subsection{Yield prediction}
ML algorithms can be used to predict crop yields by analyzing various data sources such as weather data, soil data, and satellite imagery. The specific type of ML algorithm used, and the data sources incorporated will depend on the specific crops and growing conditions being considered Table \ref{table8}.

\begin{longtable}{p{0.5in}p{0.7in}p{1.1in}p{0.8in}p{0.8in}p{1.6in}}
\caption{Overview on recent machine learning applications in yield prediction.}
\label{table8}\\ \toprule
\textbf{Ref.} & \textbf{Application} & \textbf{Feature} & \textbf{ML model} & \textbf{Performance} & \textbf{Limitation} \\ \midrule
\cite{luo_accurately_2022} & Wheat yield prediction & Satellite data.
Vegetation indices.
Soil properties.
Climate data. & RF,\newline LightGBM
\newline 
LSTM & $R^2:6.7\%$\newline
RMSE: 6.3\% & Uncertainties caused by the coarse spatial resolution of 1 km, which could affect the temporal signature of winter and spring wheat. \\ \midrule

\cite{wang_winter_2020} & Wheat yield prediction &
Climate information.\newline
Vegetation indices. & DNN &
$R^2: 0.77$,\newline
RMSE:721 kg/ha\newline
MAPE: 15.38\% & Yield forecast could be achieved at least one month before harvest with a loss of accuracy. \\ \midrule

\cite{cai_integrating_2019} & Wheat yield prediction & Crop land information.\newline
Crop yield data.\newline
Satellite based SIF.\newline
Climate information.\newline
Vegetation indices. & LASSO,\newline RF,\newline SVM,\newline NN	& $R^2: 0.75$ & Static wheat growing area. This can conduct errors when extracting the area for satellite and climate data in modeling yield and errors when aggregated yield is needed.\newline
Additional information, such as soil property information and satellite data from other spectral bands, must be added to the model to improve crop yield prediction. \\ \midrule

\cite{guo_crop_2014} & Wheat yield prediction & Crop land information. & CNN & MSE: 0.0001 & Challenges arise when training and overfitting can occur, which reduces its ability to generalize. \\ \midrule

\cite{pantazi_wheat_2016} & Wheat yield prediction & Multi-layer soil data; and satellite imagery. & SNK model & Accuracy 81.65\% & Inability to model continuous output relations. \newline
Appropriate for classification tasks only.  \\ \midrule

\cite{cao_identifying_2020} & Wheat yield prediction & Crop yield data. Cropland\newline information.
Soil's characteristics.
Climate information.\newline
Vegetation indices.\newline
Socio-economic factors. & RR, RF, LightGBM) & $R^2:$ 0.68~0.75 & They must produce more accurate crop classification yearly to reduce potential errors. Also, they only used optical and near-infrared vegetation indices (VIs). \\ \midrule

\cite{cao_identifying_2020} & Wheat yield prediction & Climate information.\newline
Soil's characteristics.\newline
Vegetation indices.\newline
Crop land information.\newline
Crop yield data. & GPR, SVM, RF. &
$R^2$\newline
GPR: 0.79\newline
SVM: 0.77\newline
RF: 0.81 & Low resolution for used remote sensing, meteorological and soil datasets, which make the predicted yields potentially uncertain. \newline
Lack of new variables for yield prediction such as SIF and radar. \\ \midrule

\cite{bhojani_wheat_2020} & Wheat yield prediction & Climate information & ANN(MLP) & MAE, MSE, RAE, RMSE, RRSE. & It is suggested to use different crop, weather, and soil datasets for crop yield forecasting at different levels. \\ \midrule

\cite{nevavuori_crop_2019} & Wheat and barley yield prediction & RGB data; NDVI data & CNN & MAPE: 8.8\% & CNN architecture performs better with RGB images than NDVI images.\newline
The model should be trained on a more extensive set of features (soil and climate) along with time series image data to adjust the trained model for accuracy. \\ \midrule

\cite{kung_accuracy_2016} & Yield prediction & Yield data.
Meteorology\newline data; Environment data;\newline and Economic data. & ENN method & Error rate: 1.30\% & Randomly creates a plurality of network for analysis. \\ \midrule

\cite{qiu_maps_2022} & Yield prediction & MODIS images\newline
Vegetation indices & Crop mapping algorithms & $R^2: 0.89$ & The quality of utilized in cropland yields is considerably low due to their high geography heterogeneity.
Generates considerable errors associated with the cropland dataset. \\ \midrule

\cite{kim_comparison_2019} & Corn yield prediction & Soybean yield prediction.	Satellite images.
Meteorological data.
Crop yield history.
Weather parameters.
Soil's characteristics. & Optimized DNN model & Correlation Coeff:\newline
Corn 0.945,\newline
Soybean 0.901 & A combination of the optimized DNN model and a spatial statistical model, such as GWR, should be investigated in a future work for greater accuracy improvement. \\ \midrule

\cite{khaki_cnn-rnn_2020} & Corn yield prediction\newline
Soybean yield prediction. & Historical data.\newline
Weather parameters.\newline
Soil's characteristics. & CNN-RNN model & RMSE:
Corn 24.10,\newline
Soybean 6.35 & Sensitive to many variables, including weather and soil. \\ \midrule

\cite{peng_assessing_2020} & Corn yield prediction\newline
Soybean yield prediction. & MODIS based VI’s.\newline
Satellite-based SIF.\newline
Weather parameters.\newline
Historical yield data & ANN, RF, Ridge, and LASSO Regression & RMSE:
Corn 18.11,\newline
Soybean 5.31 & Need to assess the performances of using different remote sensing data in crop yield prediction. \\ \midrule

\cite{shahhosseini_forecasting_2020} & Corn yield prediction & Soil's characteristics.\newline
Crop management data;\newline
Historical yield data;\newline
Climatic information. & LR
LASSO, XGBoost and LightGBM, RF, optimized weighted ensemble. & Optimized weighted ensemble
RRMSE: 9.56\% & Working more on the cross-validation procedure to generate improved out-of-bag predictions that better emulate test observations can be considered a future research direction. \\ \bottomrule
\end{longtable}

To accurately predict yields, it is important to use a combination of supervised and unsupervised learning algorithms. Supervised learning algorithms can be used to identify patterns in the data that are predictive of yield performance while unsupervised learning techniques can be used for clustering and dimensionality reduction. Additionally, domain-specific knowledge and expert insights should also be incorporated into the prediction process to better understand how different factors may affect yield outcomes. The choice of ML algorithm and data sources will depend on the specific goals and constraints associated with each individual yield prediction task.

Despite this, there are still several obstacles to entirely operating ML techniques to manage water:
\begin{itemize}
    \item Extensively high-quality data is often required for ML. Getting adequate data with high precision for water management systems might take time and effort.
    \item The existing methods may only apply to specific systems due to the exceedingly difficult conditions in water management systems, which prevents ML techniques from being widely used.
\end{itemize}
To meet the problems outlined above, future engineering and research techniques should take the following factors into consideration:
\begin{itemize}
    \item More sophisticated sensors, such as soft sensors, should be created and used to collect accurate data.
    \item More automatic and hybrid models should be generated according to agricultural needs.
\end{itemize}

\section{Internet of Things and Agriculture: Technology Use Cases for Smart Farming}
IoT is revolutionizing agriculture – from how farmers grow crops to how food is distributed. The main utilizations of IoT used in precision agriculture are:
\begin{enumerate}
    \item Soil Sampling and Mapping: Traditional soil sampling methods are time-consuming and expensive. Nevertheless, with IoT sensors, farmers can quickly and easily map their fields to get a real-time view of soil conditions. This information can be used to optimize fertilizer use, reduce water waste, and improve crop yields.
    \item Fertilization: Applying the right amount of fertilizer is crucial for healthy. With IoT sensors, farmers can monitor soil nutrients in real-time and adjust fertilization accordingly. This not only saves money on unnecessary inputs but also helps protect against environmental contamination from the overuse of chemicals. Fertilization is a key phase in closing yield gaps.
    \item Disease Monitoring: Plant diseases can spread quickly through fields, causing wide-spread damage. By monitoring plant health with IoT sensors, farmers can identify early signs of disease outbreaks so they can take action to prevent them from spreading.  
    \item Yield Measurement: Accurately measuring crop yield has traditionally been a challenge for farmers. However, with new IoT technologies, farmers can now measure yield more accurately in real time. This information allows them to adjust throughout the growing season to maximize production.
    \item Forecasting: Weather forecasting is essential for agricultural planning, but forecasts are often inaccurate or incomplete. However, by combining data from multiple sources, including weather stations, satellites. IoT sensors can provide more accurate forecasts that help farmer plan for extreme weather events such as droughts or floods.
\end{enumerate}

The rapid development of AI, the IoT, and wireless sensor networks (WSNs) is having a profound impact on modern agriculture. These technologies enable farmers to increase yields, reduce costs, and improve sustainability. AI is used to develop more efficient irrigation systems \cite{sinwar2020ai,blessy2021smart}, optimize crop production \cite{eli2019applications,oluyemi2022benefits}, and predict weather patterns \cite{al2021ai}. IoT sensors monitor water use \cite{ragab2022iot,phasinam2022application}, soil moisture levels \cite{srivastava2020monitoring}, and temperature \cite{karthikeyan2021iot}. WSNs are used to track livestock movements and grazing patterns \cite{nabi2022wireless,rahaman2022wireless}. All these data-driven innovations are helping farmers to produce more food with fewer inputs. The future of agriculture looks bright, thanks to the continued advancement of AI, IoT, and WSN technologies. Farmers who embrace these technologies will be able to position themselves for success in the 21st-century marketplace. Precision agriculture is a modern farming practice it refers to using technology to manage crops on a field-by-field basis to optimize yield and minimize inputs such as water and fertilizer. AI, IoT, and WSNs can be significantly used in precision agriculture. AI can be used for crop monitoring and yield prediction, while IoT devices for soil moisture sensing and irrigation control. WSNs to collect data from various sensors placed throughout a field. By increasing the efficiency of agricultural production, we can reduce wastage, lower costs, and increase profits while positively impacting the environment through reduced input usage.

\subsection{Valuing data for responsible agriculture}
\begin{figure}[h!]
    \centering
    \includegraphics[width=0.8\textwidth]{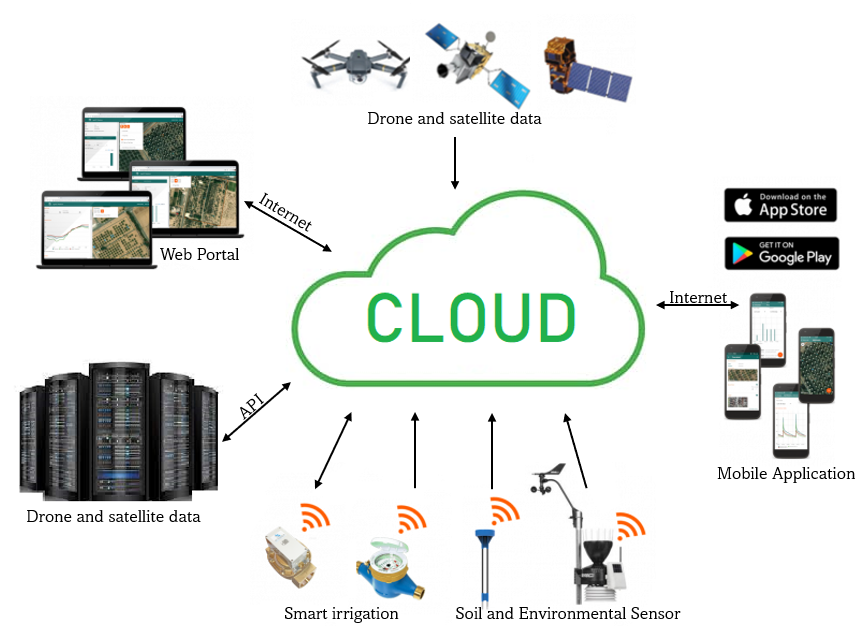}
    \caption{Agriculture cloud system. Source: Authors’ conceptualization.}
    \label{fig9}
\end{figure}

Smart agriculture is a comprehensive, sustainable, and inclusive agricultural model that promotes an equitable distribution of value along the value chain; and allows for increased production and productivity as well as improved logistics and storage capacity while improving environmental practices and resource efficiency. However, this requires capturing in real-time all relevant data produced by the stakeholders in the value chain so that their decisions are rational and efficient. Accordingly, information and communication technologies (ICT) is the apparent solution for responsible agriculture \cite{mapiye2021information} by using a IOT based wireless sensor network, cloud system, and web and mobile applications for improving crop yields, reducing resource waste, and promoting sustainable farming practices (Figure \ref{fig9}).

\subsection{Wireless sensor networks (WSNs)}
WSNs are an essential tool for precision irrigation. Soil and plant-based near-real-time monitoring sensors can give farmers the information they need to decide when and how much water to apply to their crops. This technology can help farmers reduce water use, save money, and improve crop yields \cite{rahaman2022wireless}. Precision irrigation using WSNs allows farmers to optimize their water use based on specific field conditions example \cite{hamami2020application}. The sensors collect data on soil moisture levels, temperature, rainfall, evapotranspiration rates, etc., then transmit them wirelessly to a central database. Farmers can access this information via computer or mobile application and adjust their irrigation schedule accordingly. In many cases, precision irrigation using WSNs has significantly reduced water use. In \cite{oussama2022fast} the authors developed a rapid and intelligent irrigation system based on a wireless sensor network (WSN), which resulted in a 19\% reduction in water use. In addition to saving water resources, precision irrigation can also save farmers money on their energy invoices and reduce the damage to equipment. Moreover, it can improve crop yields by preventing over- or under-watering. WSNs have been classified in various ways. One standard classification is based on the mobility of sensor nodes, which can be either static or mobile \cite{zhang2012review}. Another classification is based on the deterministic of node behavior \cite{priyadarshi2020deployment}, which can be either deterministic or nondeterministic. Static WSNs are those in which sensor nodes are deployed in a fixed position and do not change their location over time. Static WSNs are typically used for applications such as monitoring environmental conditions or tracking objects. Mobile WSNs may be used for similar applications, but they also offer the potential for new applications, such as battle-field surveillance, search-and-rescue operations \cite{agrawal2018detection}. 

\subsection{Cloud-Based Smart Farming Applications Based on Machine Learning and IoT}
Cloud computing refers to using internet-based servers and data centers to host hardware and software rather than on local devices. Arduino and Raspberry Pi are microcon-troller boards that are often used in IoT systems \cite{guntaka2022overview}. They can be integrated with cloud platforms to take edge of the scalability and flexibility of the cloud. The benefits of using cloud computing for IoT projects include cost savings, increased competitiveness, data backup and recovery, improved accuracy, and fast deployment of devices to the cloud. However, there are also potential weaknesses to consider, such as the risk of hardware or software failure, performance variability, cloud stability risk, downtime, lack of support, and the need for internet connectivity. Some examples of IoT applications that can be useful for cloud based IoT projects Table \ref{table9} include data analysis, asset tracking, and remote monitoring.

\begin{table}[h!]
\centering
\caption{Most used platforms for Internet of Things (IoT) projects.}
\label{table9}
\begin{tabular}{lccccc} \toprule
\textbf{Platforms} & \begin{tabular}{@{}c@{}}
    \textbf{Service}\\
    \textbf{integration} 
\end{tabular} & \begin{tabular}{@{}c@{}}
    \textbf{Data} \\
    \textbf{Storage}
\end{tabular} & \begin{tabular}{@{}c@{}}
    \textbf{Data}\\
    \textbf{Visualization}
\end{tabular} & \begin{tabular}{@{}c@{}}
    \textbf{SDK /}\\
    \textbf{APK}
\end{tabular} & \textbf{Cost} \\ \midrule
Thing Speak & \checkmark & \checkmark & \checkmark & \checkmark & Free \\ \midrule
AWS & $\times$ & \checkmark & $\times$ & \checkmark & \$ \\ \midrule 
Microsoft Azure & \checkmark & \checkmark & \checkmark & \checkmark & \$ \\ \midrule
Firebase & \checkmark & \checkmark & $\times$ & \checkmark & Free \\ \midrule
Kaa & \checkmark & \checkmark & \checkmark & \checkmark & Free \\ \bottomrule
\end{tabular}
\end{table}

Table \ref{table9} lists various parameters related to cloud platforms that can be used to integrate with IoT devices. These parameters include:

\begin{itemize}
    \item Services integration: tools and technology that are used to connect different applications and systems. Service integration may include the ability to send notifications (Email, SMS).
    \item Data store: describes whether the cloud platform provides storage capabilities.
    \item Data visualization: refers to the ability to represent information in the form of graphs, charts, and images.
    \item SDK and API: A Software Development Kit (SDK) is a toolkit used to develop software applications, while an Application Programming Interface (API) allows software programs to communicate with each other.
    \item Free account: This indicates whether the cloud platform offers a free account option.
\end{itemize}

There are several ways that a cloud system could be used to support smart farming applications based on ML and the IoT \cite{abu2022internet,farooq2019survey,javaid2022enhancing,zamora2019smart,muangprathub2019iot,khattab2016design}:

\begin{itemize}
    \item Sensor data collection and analysis: IoT sensors and devices can be used to collect data on various aspects of a farming operation, such as soil moisture, crop health, and weather conditions. This data can be transmitted to a cloud system, where ML algorithms can be used to analyze it and generate insights and predictions.
    \item Predictive maintenance: used to predict when certain farming equipment or infrastructure is likely to fail, allowing farmers to proactively schedule maintenance and avoid downtime.
    \item Crop yield prediction: ML algorithms can be used to analyze data from sensors and other sources to predict crop yields, helping farmers to optimize their operations and reduce waste.
    \item Water management: IoT sensors and devices can be used to monitor soil moisture levels and water usage, allowing farmers to optimize irrigation and conserve water resources.
    \item Pest and disease management: ML algorithms can be used to analyze data from sensors and other sources to identify pests and diseases in crops, allowing farmers to take early action to prevent damage.
    \item Supply chain optimization: ML algorithms can be used to analyze data from sensors and other sources to optimize the supply chain, reducing waste and improving efficiency.
\end{itemize}

Overall, a cloud system can be a powerful platform for supporting smart farming applications based on ML and IoT, helping farmers to make more informed decisions and improve the efficiency and sustainability of their operations.

\subsection{Platforms, Applications}
The platform using ML and the IoT could be developed to help farmers predict and prepare for various aspects of crop growth and resource management. The application use ML algorithms to analyze data from sensors, satellite imagery, and other sources to predict crop yields, irrigation needs, soil health, and other factors. The data could be collected and transmitted via the IoT, using a cloud platform to store and process the data \cite{ojha2021internet}. The web application provides farmers with real-time information and recommendations to help them optimize their operations and make more informed decisions. For example, the application could send alerts when irrigation is needed or when certain crops are at risk of pests or diseases. It could also provide information on the best times to plant and harvest different crops and suggest methods for improving soil health and reducing resource waste. Many applications provide various tools for managing farm operations, from planting to harvesting crops Table \ref{table10}. They offer data-driven insights into soil conditions, weather forecasts, pest control strategies, and more so that users can make informed decisions about how to run their farms efficiently. Furthermore, some applications have an integrated marketplace where users can buy or sell products related to farming, such as seeds or fertilizers, directly from other producers at competitive prices. 

\begin{table}[h!]
\centering
\caption{Example of the highest-ranking applications for agriculture monitoring and forecasting.}
\label{table10}
\begin{tabular}{lp{0.5in}p{2.35in}ccc} \toprule
\textbf{Ref.} & \textbf{App Name} & \textbf{Application} & \textbf{Android} & \textbf{IOS} & \textbf{Webpage} \\ \midrule
\cite{cao_igrow_2022} & Tencent’s iGrow & AI planting solution, helps reduce labor costs by 20-25\% and heating costs by 30-40\%. & \checkmark & \checkmark & \checkmark \\ \midrule
\cite{farmLogs2022} & FarmLogs & Managing farms from planting to harvesting crops & \checkmark & \checkmark & \checkmark \\ \midrule

\cite{janssen_online_2009} & AgrarWetter & Estimation of weather and evaporation to guide irrigation decisions. & \checkmark & \checkmark & \checkmark \\ \midrule

\cite{manimala_agriapp_2020} & AgriApp & Production and crop management. & \checkmark & $\times$ & \checkmark \\ \midrule

\cite{fan_case_2022} & Pinduoduo & Promotes smart Agri-tech using advanced agricultural technologies and devices. & \checkmark & $\times$ & \checkmark \\ \midrule

\cite{gan_development_2018} & Smart Farm & Remote monitoring and control app for crop management. & \checkmark & \checkmark & $\times$ \\ \midrule

\cite{anurag_agro-sense_2008} & AgroSense & Soil Health tracking, overall planning, and budgeting & \checkmark & \checkmark & \checkmark \\ \midrule

\cite{kassim_iot_2020} & FarmOS & Crop management, labor management, order management. & $\times$ & $\times$ & \checkmark \\ \midrule

\cite{hayes_vineyard_2020} & Onesoil & Remote crops monitoring, increase yields, and reduce seed and fertilizer costs & \checkmark & \checkmark & \checkmark \\ \bottomrule
\end{tabular}
\end{table}

Overall, these mobile apps have been making life easier for farmers everywhere by automating tedious tasks associated traditionally done manually, giving them access valuable data points easily accessible. Farmlogs and Fieldview are examples of services that uses AI algorithms and radar data from the National Oceanic and Atmospheric Administration (NOAA) to estimate rain intensity, that creates an option to replace a rain gauge unavailable to the farmer \cite{henry2020arkansas}. In summary, this type of data-driven approach can improve crop yields, reduce resource waste, and promote sustainable farming practices, ultimately contributing to the goal of responsible agriculture.

\section{Overview of key observations}
To ensure high output, an effective and updated crop monitoring system is required. Several information gaps still need to be covered to improve this system's accuracy. For instance, additional research is required to determine how various algorithms might be implemented for yield prediction. This section presents an overview of key observations, including statistics, critical review, challenges, and recommendations of using AI, ML and IoT in agriculture, is a comprehensive summary of a particular research study or project's main findings, strengths, and limitations. 

\subsection{Statistics}
\begin{figure}[h!]
    \centering
    \includegraphics[width=0.75\textwidth]{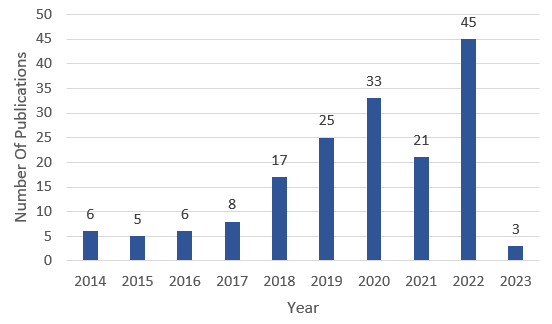}
    \caption{Most Recent publications used in this manuscript.}
    \label{fig10}
\end{figure}

The most relevant papers on the use of machine learning in predicting agriculture were selected (Figure \ref{fig10}). The analysis focused on the most common data features used, including satellites, weather and soil data (Table \ref{table11}), as well as the most frequently used ML algorithms (Table \ref{table12}) and evaluation metrics (Table \ref{table13}). This analysis provides a comprehensive overview of the current state-of-the-art in the field of ML in predicting agriculture.

\begin{table}[h!]
\centering
\caption{Most used Features in forecasting agriculture.}
\label{table11}
\begin{tabular}{lcc} \toprule
\textbf{Features} & \textbf{Number of usages} & \\ \midrule
Satellite data & 30 & \multirow{5}{*}{\includegraphics[width=2.2in]{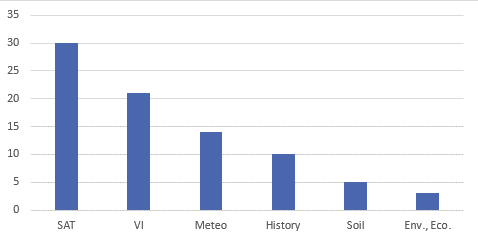}} \\  \cmidrule{1-2}
Vegetation index & 21 \\ \cmidrule{1-2}
Meteorological data & 14 \\ \cmidrule{1-2}
Historical yield data & 10 \\ \cmidrule{1-2} 
Soil’s characteristics & 5 \\ \cmidrule{1-2}
Environment and economic data & 3 \\ \bottomrule
\end{tabular}
\end{table}

\begin{table}[h!]
\centering
\caption{Distribution of the widely used machine learning models in forecasting agriculture.}
\label{table12}
\begin{tabular}{lcc} \toprule
\textbf{Algorithm} & \textbf{Number of usages} & \\ \midrule
SVM & 13 & \multirow{5}{*}{\includegraphics[width=2.25in]{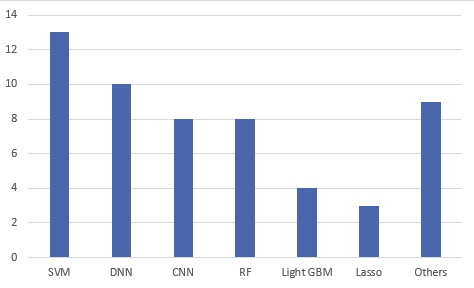}} \\  \cmidrule{1-2}
DNN & 10 \\ \cmidrule{1-2}
CNN & 8 \\ \cmidrule{1-2}
RF & 8 \\ \cmidrule{1-2}
Light GBM & 4 \\ \cmidrule{1-2}
Lasso & 3 \\ \cmidrule{1-2}
Others & 9 \\ \bottomrule
\end{tabular}
\end{table}

\begin{table}[h!]
\centering
\caption{Most used evaluation metrics used in forecasting agriculture.}
\label{table13}
\begin{tabular}{lcc} \toprule
\textbf{Features} & \textbf{Number of usages} & \\ \midrule
MSE & 29 & \multirow{5}{*}{\includegraphics[width=2.25in]{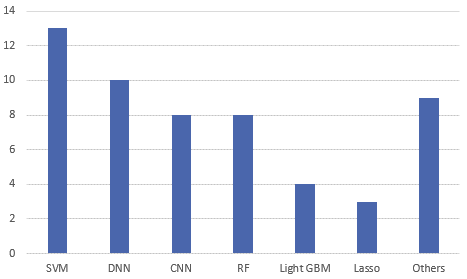}} \\  \cmidrule{1-2}
Accuracy & 25 \\ \cmidrule{1-2}
RMSE & 19 \\ \cmidrule{1-2}
$R^2$ & 9 \\ \cmidrule{1-2}
R & 8 \\ \cmidrule{1-2}
Correlation & 4 \\ \cmidrule{1-2}
Others & 6 \\ \bottomrule
\end{tabular}
\end{table}

\subsection{Critical review}
One of the main benefits of using ML and IoT in agriculture is the ability to collect and analyze large amounts of data from a variety of sources, including sensors, weather stations, and satellite imagery \cite{benos2021machine}. This data can be used to generate insights and predictions that can help farmers make more informed decisions, such as when to plant, irrigate, and harvest crops. However, there are also a few challenges and limitations to the use of ML and IoT in agriculture. One of the main challenges is collecting and processing massive high-quality data, which is essential for training accurate ML models \cite{roh2019survey}. In many cases, data from sensors and other IoT devices may be noisy or incomplete, which can make it difficult to generate reliable insights and predictions. Although most data collection approaches involve collecting data and then using it to train a model, another critical approach is to augment or improve the data based on how the model performs. This involves using the model's output or other metrics to identify areas where the data may be insufficient or inaccurate and then collecting additional data or refining the existing data to address these issues. This can be an essential step in improving the performance and reliability of the model and can help to ensure that it is based on the most accurate and relevant data possible. Another challenge is ensuring that the necessary infrastructure is in place to support the use of ML and IoT in agriculture. This includes things like internet connectivity, power supplies, and hardware. In addition, building and implementing ML models and IoT systems can require specialized knowledge and resources that may not be readily available in the agricultural sector.  Also, the costs associated with implementing ML and IoT technologies in agriculture can be high, especially for small farmers in developing countries. Additionally, it may be difficult to scale these technologies to larger operations or across regions. A critical review and evaluation of recent review articles in the areas of ML in Table \ref{table14} involve examining a range of research studies and reports to identify common themes, trends, and gaps in the current knowledge base.

\begin{table}[h!]
\centering
\caption{Critical evaluation of the recent review articles in machine learning and forecasting agriculture.}
\label{table14}
\begin{tabular}{p{1.5in}p{2.1in}p{2.1in}} \toprule
\textbf{Ref.} & \textbf{Objectives of the review} & \textbf{Limitation of the review} \\ \midrule
\cite{chong2017review} & Remote sensing for plant monitoring: gaps identification and recommendation for future research. & This review contains a lack of potential emphasis on yield prediction algorithms and performances evaluation. \\ \midrule
\cite{chlingaryan2018machine} & Reviews ML for crop yield prediction and nitrogen status estimation using Hybrid systems and signal processing. & Analysis of features and performance,\newline
Critical evaluation of existing ML for crop yield prediction. \\ \midrule 

\cite{rashid_comprehensive_2021} & Reviews the use of ML algorithms to predict palm oil yield prediction. & The existing palm oil yield prediction listed studies utilized limited number of features which results in a significant difference between the predicted and actual palm oil yield. Thus, more research should be done using several features and a variety of prediction methods and recommendations for future directions. \\ \midrule
\cite{van2020crop} & A variety of features, ML algorithms and metrics have been reported & Lack of: Critical evaluation, recommendation, and future directions. \\ \midrule

\cite{bali_emerging_2022} & Evaluation of several ML methods used in crop yield prediction. & Need to explore more hybrid ML models.\newline
Lack of recommendations. \\ \midrule

\cite{elavarasan_forecasting_2018} & Overview of the existing crop yield ML models. Compare approaches using different performance & All current crop yield prediction models are based entirely on climatic conditions; other characteristics must be considered for yield prediction. \\ \midrule

\cite{woittiez2017yield} & Overview of the existing data on yield limiting factors and current causes of yield gaps & Does not evaluate any ML approach for closing the yield gap, and lack of recommendation and perspectives for future research. \\ \midrule

\cite{ahansal2022towards} & Utilization of Geospatial Technologies and ML for Water Resource Management in Arboriculture & Lack of actual case studies in using ML and IoT-based UAV sensors when comparing the most popular UAV sensors and vegetation indices. \\ \midrule
\cite{abioye2022precision} & Presents an overview of the existing ML models in precision irrigation and water management & Lack of performance evaluation for water management activities,\newline
Need to explore: Hybrid and automatic ML algorithms,\newline
And platforms used for IoT based ML projects for precision irrigation. \\ \midrule

\cite{saleem2021automation} & Presents robotic solutions for the major agricultural tasks by machine and deep learning algorithms & This review lacks critical evaluation of existing ML models used for the major agricultural tasks.\newline
Identify factors impacting the performance of ML/DL algorithms for agricultural activities. \\ \bottomrule
\end{tabular}
\end{table}

\clearpage

Some key points that might be addressed in a review could include the following:
\begin{itemize}
    \item The use of ML techniques for agriculture forecasting: This could involve examining various approaches and algorithms used in literature to analyze and predict various aspects of agriculture and environmental phenomena and evaluating their strengths and limitations.
    \item The role of data acquisition in ML for agriculture: examining different sources including satellites, weather data, imagery, and other types of data, and evaluating their quality and relevance to the research requirements.
    \item The performance of ML algorithms: comparing and examining multiple metrics and evaluation techniques used to assess the performance of different ML algorithms and identify their accuracy and reliability.
    \item The potential of hybrid ML models: examining hybrid models that combine multiple ML algorithms or approaches and evaluating their potential benefits and challenges.
    \item Recommendations for future directions: making recommendations for areas of research that could be valuable to explore, such as developing new ML algorithms or integrating additional data sources.
\end{itemize}

\subsection{Challenges and limitations of using machine learning and IoT in agriculture}
There are several challenges and limitations to using ML and IoT in agriculture. Some of these include:
\begin{itemize}
    \item Data collection and quality: One of the main challenges in using ML and IoT in agriculture is collecting and preprocessing high-quality data \cite{christias2021machine}. In many cases, data from sensors and other IoT devices may be noisy or incomplete, which can make it difficult to train accurate ML models.
    \item Infrastructure and connectivity: Another challenge is ensuring that the necessary infrastructure is in place to support the use of ML and IoT in agriculture. This includes things like internet connectivity, power supplies, and hardware \cite{islam2021review}. In many rural areas, these resources may be limited or unavailable, which can make it difficult to implement these technologies.
    \item Expertise and resources: Building and implementing ML models and IoT systems can require specialized knowledge and resources that may not be readily available in the agricultural sector. This can make it difficult for farmers and other stakeholders to adopt these technologies.
    \item Cost and scalability: The costs associated with implementing ML and IoT technologies in agriculture can be high, especially for small farmers or those in developing countries. Additionally, it may be difficult to scale these technologies to larger operations or across regions.
    \item Regulation and privacy: The use of ML and IoT in agriculture may also be subject to various regulations and privacy concerns. For example, data collected from IoT devices may be sensitive and may need to be protected in accordance with relevant laws and regulations.
\end{itemize}

\section{Future trends}
The application of ML in agriculture can revolutionize how we grow crops and manage our natural resources. ML can monitor irrigation systems \cite{abioye2022precision,vianny2022water}, predict crop yields \cite{bali_emerging_2022}, and optimize fertilization and irrigation schedules \cite{abioye2022precision}. Additionally, ML can be used to develop intelligent applications for monitoring and managing the agriculture sector. Thus, the usage of ML in agriculture has the potential to increase yields while reducing input costs. For example, by using ML to monitor irrigation systems, farmers can reduce water usage without surrendering crop yield \cite{truong2021machine,bhoi2021automated}. In addition, by using ML to predict crop yields, farmers can reasonably plan their production schedules and sidestep over-production or underproduction. 

\begin{figure}[h!]
    \centering
    \includegraphics[width=0.65\textwidth]{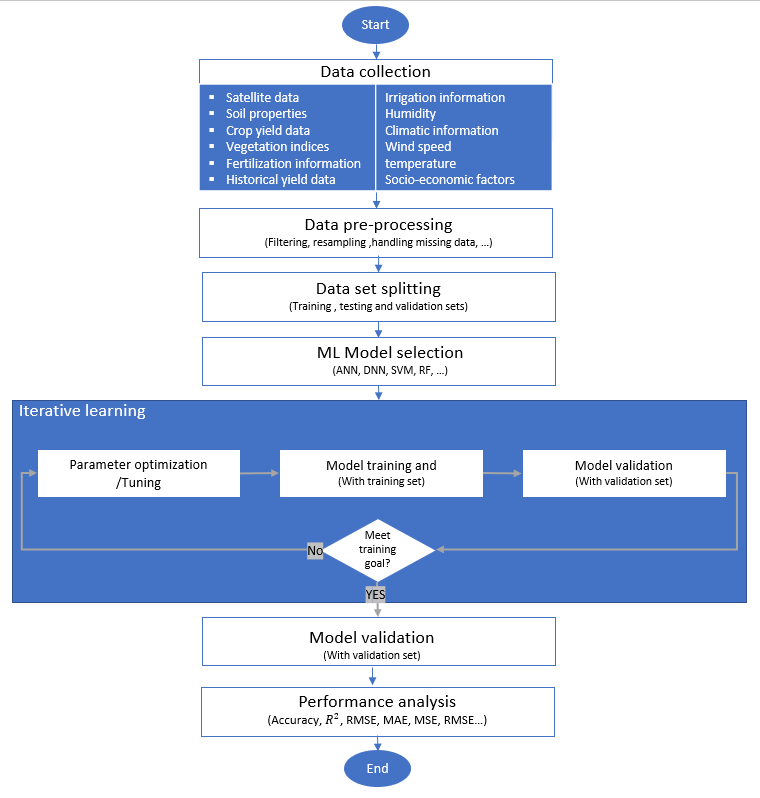}
    \caption{Automatic ML modeling and application procedure.}
    \label{fig11}
\end{figure}

In general, the process of ML for prediction is composed of several critical phases as pursues. Firstly, the available data set will be divided into training, validation, and testing sets. Data preprocessing uses specific methods of analyzing, filtering, transforming, and encoding data. A specific ML model is selected, which will be trained and validated based on the training and validation sets. Before testing with untrained data, the related hyperparameters will be tuned repeatedly until the preset training goal (precision) is met. Eventually, the testing set will be used to test the trained model and to evaluate the performance. For clarity, a flow chart of an automatic ML modeling and application procedure is given in Figure \ref{fig11}.

\clearpage

Farmers can use ML predictions to decide what crops to plant, how much fertilizer to use, and when to harvest. Many factors can affect crop yields, so it is vital to build a model that can account for many of these factors. Weather is one of the most critical factors affecting agriculture, thus data from weather stations can be used in the models. Other data sources include satellite images, soil moisture sensors, and information from agricultural databases. The models built using ML and deep learning techniques are constantly improving. They are becoming more accurate at predicting yield for specific locations and conditions. This information can help farmers reduce risk and increase profits by making better crop decisions.

Agriculture forecasting uses ML and deep learning algorithms to predict future crop yields, prices, and other agricultural trends. Agriculture forecasting can be used to decide on planting, harvesting, and marketing strategies \cite{sinha_recent_2022}. It can also help farmers and agri-businesses anticipate weather patterns and plan for natural disasters \cite{fahad_implementing_2023}. Also, ML algorithms are well-suited for agriculture forecasting because they can automatically learn from data sets that are too large or complex for humans to analyze. Deep learning algorithms can go one step further by making predictions based on multiple layers of data (e.g., weather patterns, soil conditions, etc.). This makes them particularly well-suited for long-term forecasts such as future planning purposes. There are many potential applications for agriculture forecasting, including helping farmers decide when to plant or harvest their crops, supporting agribusinesses in making pricing and product availability decisions, and informing government policies that support the agricultural sector. Moreover, ML algorithms can analyze large volumes of data from various sources, such as weather stations, satellites, and field sensors, to identify patterns and make more accurate predictions about crop yields and other factors. As more data is collected from various sources, ML models will become robust at making accurate predictions about crop yields, water management and other key activities. Automatic ML models can ingest large volumes of data and use this information to identify patterns that can be used to make more accurate predictions to meet specific agricultural goals \cite{kundu2021iot}, thus increased focus on improving accuracy and processing time is required. The combination of ML and IoT can provide real-time updates and alerts about potential adverse conditions (e.g., drought, early frost, flood, ...). This allows farmers to take proactive measures to protect their crops and minimize losses.

\section{Recommendations}
Predicting future crop yields and performance based on various data sources is challenging \cite{martos2021ensuring}. The recommended approach for crop yield prediction is to use historical and weather data in combination with Sentinel 2 satellite data and VI’s \cite{bukowiecki_sentinel-2_2021,saleem2021automation,christias2021machine,islam2021review}. This method offers temporal and spatial resolution advantages, allowing for better observation of water resources, crop growth, vegetation indices, and estimation of crop yield. Utilizing a combination of historical data, weather conditions, and satellite imagery along with machine learning algorithms especially XGBoost, Light GBM, Gradiant Boost, ANN, SVM and Random Forest has shown the best results according to recent studies. Combining these approaches and develop a hybrid ML model will take the advantages of each model and gives more advantageous results. Also, recommendation of the use of multiple metrics, mainly R$^2$, RMSE, and overall accuracy score, those metrics validate the accuracy of the predictions and identify potential issues before deployment, also used to compare different models to decide which model works better on a particular set of circumstances or parameters. Furthermore, these metrics helps identify any potential issues with the model before deployment.

\begin{figure}[h!]
    \centering
    \includegraphics[width=0.75\textwidth]{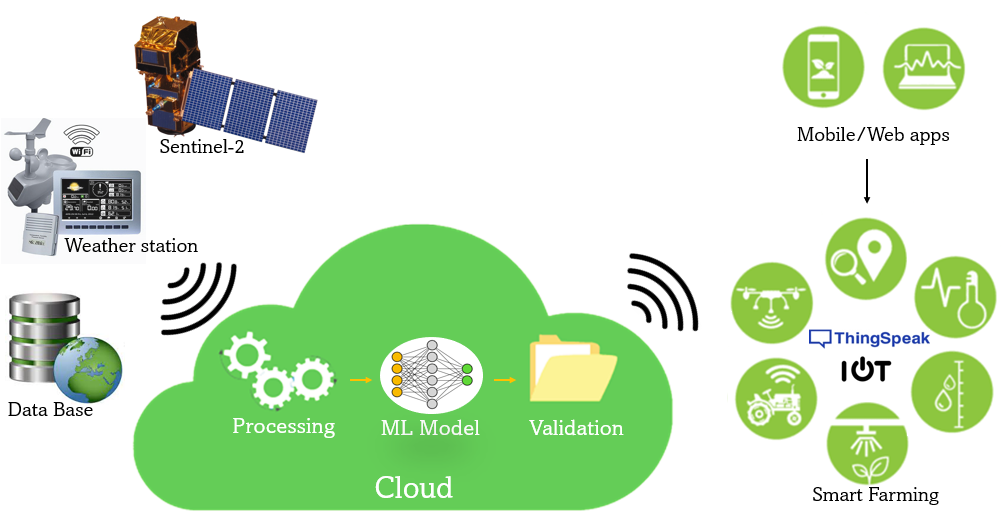}
    \caption{Recommendation for a complete agricultural system. Source: Authors’ conceptualization.}
    \label{fig12}
\end{figure}

Overall, developing digital software applications designed explicitly around ML-based solutions and IoT (Figure \ref{fig12}) requires significant investment into research and development efforts which may prove too costly for smallholder farmers who need more resources to support these initiatives financially. Fortunately, there are several ways that these obstacles can be addressed through the cloud system by providing open access platforms where users can upload their own data sets, share insights, and collaborate on research projects this could potentially help bridge the gap between those who have access when it comes to leveraging this powerful toolset. This would also provide an opportunity to build more customized solutions tailored to meet specific needs based on user feedback and inputs.

\section{Conclusion}
The lack of access to high-quality data sets for training models and the high initial cost of deployment pose significant obstacles for small-scale farmers. However, the use of open access platforms where users can share data sets and collaborate on research projects can help address these challenges and provide more customized solutions tailored to meet specific needs. The importance of collaboration and partnerships between farmers, researchers, and industry stakeholders in ensuring the widespread adoption of ML-based solutions in the agriculture sector cannot be overstated. Further research could focus on addressing the challenges of limited accessibility and reducing the initial cost of deployment for small-scale farmers. In conclusion, this work highlights the value of using multiple data sources and advanced ML and IoT techniques to make more accurate predictions and real-time applications in the field of agricultural forecasting. One weakness of this work is that it does not provide specific details on how to overcome the challenges of limited accessibility and the high initial cost of deployment. To address this weakness, further research could focus on identifying practical solutions for increasing accessibility and reducing costs for small-scale farmers. One strength of this work is its emphasis on the importance of collaboration and partnerships between farmers, researchers, and industry stakeholders in ensuring the widespread adoption of ML-based solutions in the agriculture sector. In conclusion this work is valuable in the domain of agricultural forecasting as it highlights the importance of using multiple data sources and advanced ML and IoT techniques to make more accurate predictions and real time applications. It also emphasizes the need to consider the initial cost of deployment and the challenges that small-scale farmers may face in adopting these technologies.

\bibliographystyle{unsrtnat}
\bibliography{references}  






\end{document}